\documentclass{article}

\usepackage{microtype}
\usepackage{graphicx}
\usepackage{subfigure}
\usepackage{booktabs} 

\usepackage{hyperref}

\usepackage[accepted]{icml2024}

\usepackage{amsmath}
\usepackage{amssymb}
\usepackage{mathtools}
\usepackage{amsthm}

\usepackage[capitalize,noabbrev]{cleveref}

\usepackage{placeins} 
\usepackage{longtable} 

\theoremstyle{plain}

\theoremstyle{definition}

\theoremstyle{remark}

\usepackage[textsize=tiny]{todonotes}

\icmltitlerunning{Submission and Formatting Instructions for LCFM 2024}

\begin{document}

\twocolumn[
\icmltitle{Unable to Forget: Proactive Interference Reveals Working Memory Limits in LLMs Beyond Context Length}
\vspace{-0.4cm}


\icmlsetsymbol{equal}{*}

\begin{icmlauthorlist}
\icmlauthor{Chupei Wang}{equal,uva}
\icmlauthor{Jiaqiu Vince Sun}{equal,nyu}

    \normalsize 
    \textsuperscript{1}University of Virginia     
    \textsuperscript{2}New York University-Center for Neuroscience\\
    {\textsuperscript{1}cw4bb@virginia.edu \textsuperscript{2}vince.sun@nyu.edu}

\end{icmlauthorlist}




\vskip 0.25in
]




\renewcommand{\thefootnote}{\fnsymbol{footnote}}
\footnotetext[1]{Equal contribution. Listing order is random. Details in Contributions section.}
\renewcommand{\thefootnote}{\arabic{footnote}}

\begin{abstract}
Information retrieval in Large Language Models (LLMs) is increasingly recognized as intertwined with generation capabilities rather than mere lookup. While longer contexts are often assumed to improve retrieval, the effects of intra-context interference remain understudied. To address this, we adapt the proactive interference (PI) paradigm from cognitive science, where earlier information disrupts recall of newer updates. In humans, susceptibility to such interference is inversely linked to working memory capacity. We introduce PI-LLM, an evaluation that sequentially streams semantically related key–value updates and queries only the final values. Although these final values are clearly positioned just before the query, LLM retrieval accuracy declines log-linearly toward zero as interference accumulates; errors arise from retrieving previously overwritten values. Attempts to mitigate interference via prompt engineering (e.g., instructing models to ignore earlier input) yield limited success. These findings reveal a fundamental constraint on LLMs’ ability to disentangle interference and flexibly manipulate information, suggesting a working memory bottleneck beyond mere context access. This calls for approaches that strengthen models’ ability to suppress irrelevant content during retrieval. Code and data are publicly available at
\href{https://github.com/zhuangziGiantfish/Unable-to-Forget}{github.com/zhuangziGiantfish/Unable-to-Forget}.
\end{abstract}

\section{Introduction}
\label{intro}
\begin{figure}[h]
  \centering
  \includegraphics[width=1\linewidth]{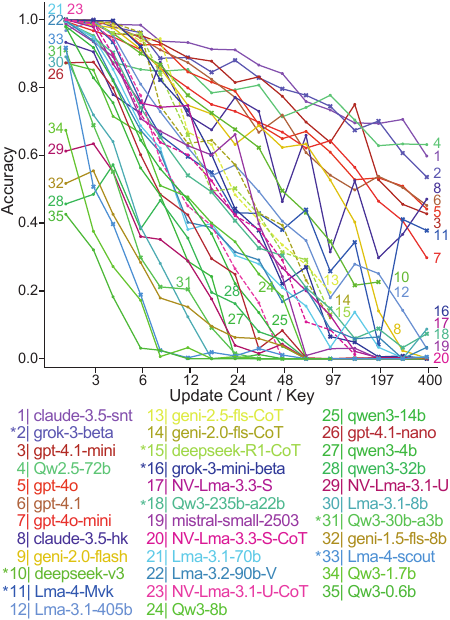} 
  \caption{Universal \textbf{log-linear} decline in retrieval performance due to interference. Increasing the amount of interfering information preceding a retrieval target within a language model’s input context results in a log-linear decrease in retrieval accuracy across diverse models. The target is positioned after the interfering information and explicitly referenced in the prompt to reduce search difficulty and isolate interference effects. (x-axis: the number of semantically similar items, log-scaled; asterisk: MoE models)}
  \label{fig:acc_demo}
\end{figure}
Evaluation for information retrieval in Large Language Models (LLMs) primarily uses input length as the main measure of task difficulty. Current research indicates that language models generally struggle with retrieval tasks when closely related pieces of information are present \citep{vodrahalli-olszewska2024MichelangeloLongContext}. Furthermore, reasoning models do not effectively improve performance in these scenarios \citep{openai_gpt41_2025}.

Current studies often conflate search difficulty—the challenge of locating the relevant “needle” in a vast contextual haystack—with interference—the challenge of correctly identifying that needle when it is surrounded by similar-looking but incorrect items. Recent long-context benchmarks—most of which evolve from the original Needle-in-a-Haystack paradigm, such as DeepMind’s Michelangelo \citep{vodrahalli-olszewska2024MichelangeloLongContext} and OpenAI’s MRCR \citep{openai_mrcr_2025} primarily raise task difficulty by lengthening the prompt. Although these studies acknowledge interfering information's impact on the retrieval tasks, they do so only in a preliminary way, without explicitly isolating or quantifying interference's independent effect on LLMs' context usage. Consequently, current research implicitly attributes the difficulty of distinguishing similar information mainly to greater input length, thereby overlooking interference as a separate, quantifiable factor.

Our work demonstrates that the amount of interfering information—semantically similar distractors—independently and significantly impacts retrieval accuracy in LLMs (Figure~\ref{fig:acc_demo}). By systematically varying interference load, we obtain the first quantitative curve that isolates interference as an independent factor. To demonstrate that interference effects are independent of input length, we include a control condition in which input length is held constant. Anti-interference capacity varies sharply across models, making it a useful discriminative trait. Crucially, even modest distractor loads expose a fundamental weakness: current LLMs cannot reliably suppress competing cues.

Interfering information consists of semantically similar text and is common in many data processing tasks. One of the simplest forms involves key–value pairs, where the key remains the same but the associated value is repeatedly updated within a sequence. For example, consider a sequence of blood-pressure (BP) readings, where the task is to keep track of the most recent BP value. BP: 120 – triage; BP: 128 – 10 min later; BP: 125 – discharge. In this task, the desired output is ‘BP: 125,’ the last-presented key–value pair. However, retrieval may be impaired by prior semantically similar BP values, which act as distractors. The search difficulty in such key-value tracking tasks is minimized, as the target answer is always the last value of a certain key. 

Notably, humans demonstrate high accuracy on these tasks. In contrast, our experiments show that \textbf{retrieval accuracy in state-of-the-art LLMs declines in a log-linear fashion as the amount of interference information preceding the target key–value pair increases}, as shown in (Figure~\ref{fig:acc_demo}), a pattern we observed consistently across all models tested.

While standard synthetic key–value retrieval tasks are widely used in LLM evaluations (e.g., Lost in the Middle), our approach uniquely leverages insights from the proactive interference (PI) paradigm in cognitive psychology. In classic PI experiments, participants recall the most recent association for a repeated cue while earlier associations cause interference. By adapting this principle for LLM testing, we fix the retrieval target as the last-presented value of a particular key, thereby minimizing search difficulty and isolating interference as an independent factor. In our experimental design, the LLM is explicitly prompted to retrieve the most recent key–value pair for a given key—ensuring the retrieval target is always the last-presented value, which minimizes search difficulty. We systematically manipulate the amount of semantically similar interfering information preceding the target and measure the effect on retrieval accuracy. This approach allows us to directly quantify the impact of interference strength, independent of search difficulty.

The experiment further demonstrated that the load of interference information induces a log-scale reduction in retrieval accuracy even when input length remains constant, revealing that input length and interference are independent factors affecting retrieval. Additionally, we observed similar log-linear declines in retrieval accuracy when manipulating other dimensions of information load, such as increasing the token length of the retrieval target. These findings indicate that LLM retrieval is constrained by a unified limit on interference tolerance, irrespective of how the interfering information is presented. This bottleneck mirrors the unitary working memory limit found in humans.

This gap underscores that, despite the task’s simplicity for humans, LLMs remain vulnerable to interference. Cognitive science research on proactive interference (PI) shows that, although humans are also affected by prior interference information, their recall performance typically plateaus: after a certain threshold, further interference produces minimal additional impairment \citep{oberauer-vockenberg2009UpdatingWorkingMemory}. This robustness is attributed to humans’ ability to actively unbind outdated associations from working memory before encoding new information \citep{oberauer-vockenberg2009UpdatingWorkingMemory}. In contrast, we hypothesize that LLMs lack such unbinding mechanisms, which may explain the continuous, monotonic decline in their retrieval accuracy as interference increases—eventually resulting in complete retrieval failure under high interference.

To disentangle the factors governing resistance to interference in large language models, we introduced the Interference Endurance Score (IES) to quantify a model’s ability to resist proactive interference. Statistical analyses across a broad range of LLMs show that larger models (i.e., with higher parameter counts) achieve substantially higher IES, whereas the nominal length of the context window has no significant effect. In other words, an LLM’s resilience to interference depends primarily on its overall representational capacity, not on how many tokens it can process.

Building on these findings, we further investigated whether LLMs could adopt human-like strategies for managing interference through explicit modulation of memory content. Humans benefit  from direct instructions to deprioritize prior interfering information \citep{festini-reuter-lorenz2014CognitiveControlFamiliarity}. To test whether similar explicit strategies could aid LLM performance, we provided natural language annotations marking the majority of prior information as outdated and irrelevant.  Despite clear instructional cues and explicit annotations, we observed only minimal improvements in LLM retrieval accuracy. In contrast to humans, who successfully utilize instructions to deprioritize or unbind outdated associations, LLMs exhibited no meaningful capacity for selective modulation of their retrieval processes under interference. This persistent limitation underscores a fundamental gap between human and LLM working memory mechanisms, highlighting a critical area for future research: developing architectures capable of flexibly adjusting their memory representations and retrieval behavior in response to instructional cues.

Based on these findings, we propose a framework of \textbf{“Limited Anti-Interference Capacity”} in LLMs, drawing a conceptual parallel with human working memory. Specifically, we argue that an LLM’s working memory–like capacity is not reflected by its input context length, but rather by an independent limit defined by its resistance to interference. Once this capacity is taxed by interfering information, retrieval performance degrades inescapably.

Our findings can be distilled into the following points.
\begin{itemize}
    \item \textbf{Interference overrides recency and instruction.}  
    Interfering information consistently and substantially degrades LLMs’ retrieval performance. Errors are dominated by retrieving prior interfering values—even when the correct answer is clearly cued and appears near the end of the input, a design that reduces search difficulty and favors retrieval based on recency. This demonstrates that neither recency bias nor explicit prompt cues are sufficient to overcome interference. LLMs are unable to “forget” or “ignore” irrelevant prior updates.
    \item \textbf{Universal Log-Linear Decay Reveals Unified Interference Limit.} 
    Across all state-of-the-art models, retrieval accuracy declines log-linearly toward zero as update count, number of tracked keys, or value length increases (Figures~\ref{fig:acc_demo},\ref{fig:acc_keys_reduced},\ref{fig:acc_value_length}). This indicates a robust, negative log-linear relationship between retrieval performance and interference load and points to a unified capacity limit for interference tolerance.
    \item \textbf{Marginal effectiveness of natural language prompt interventions.}Attempts to mitigate interference using inline natural language prompts—such as “forget” irrelevant prior information or to “focus” on the most relevant updates—yielded only marginal improvements. This underscores the robustness of interference effects and the limits of prompt engineering.
\end{itemize}

\noindent\textbf{Experimental Overview} \\
Our experiments systematically demonstrate that interference is the principal factor limiting retrieval accuracy in LLMs. \textbf{Experiment 1} reveals a robust, log-linear decline in accuracy as interference increases. \textbf{Experiment 2}, which holds input length constant, confirms that this effect is driven by interference itself rather than input length. \textbf{Experiment 3} further shows that retrieval performance is universally constrained by a single interference capacity limit, regardless of how interference is introduced. \textbf{Finally}, we investigate mitigation strategies, offering new insights into LLMs’ ability to manage in-context information under interference.

\section{Interference Dominates Retrieval Despite Recency and Instructions:}

Our objective is to understand how Large Language Models (LLMs) manage interference when retrieving information. To reduce searching difficulty and measure the impact of interference, we designed a synthetic key-value retrieval experiment.

In this test, the input is a sequence of key–value pairs, where a fixed set of keys—each representing a variable of interest—appears repeatedly throughout the sequence, each time paired with a different value. Updates for different keys are randomly interleaved, with the constraint that the same key does not appear twice in succession. This design mimics, in a simplified manner, real-world logging systems that track multiple physiological variables over time—for example, blood pressure, heart rate, and oxygen level readings recorded in a patient's health log.

After the entire sequence is presented, the language model is prompted to return the most recent value associated with each key—that is, the value from the last occurrence of each key in the sequence. All prior occurrences of the same key (with earlier values) serve as distractors. Since the retrieval target is always the value from the last occurrence of each key, the search is straightforward, and errors can be attributed primarily to interference rather than search difficulty.

\subsection{Experimental Design}

\begin{figure}[h]
  \centering
  \includegraphics[width=1\linewidth]{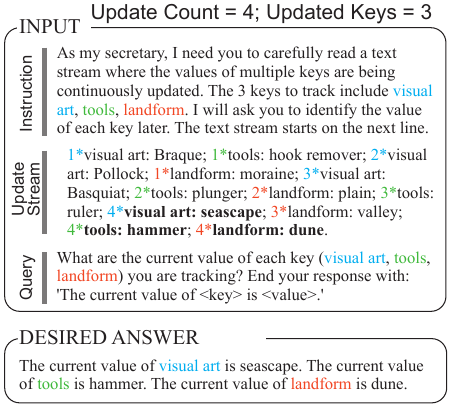}
  \caption{Basic input example for the LLM-PI test. A continuous stream of key-value updates is presented to the model, with up to 46 updated keys and 400 distinct values (update count) used in the actual experiments. In this example, three distinct keys ("visual art", "tools", and "landform")—color-coded for clarity—each undergo four updates. Numerical prefixes (e.g., "1*") denote update order for visualization purposes only and were not part of the input. The model is instructed to retrieve the final value for each tracked key, indicated in bold for illustration. The keys to check are cued both before and after the update stream.}
  \label{fig:input_updates}
\end{figure}

In this task (Figure~\ref{fig:input_updates}), each input sequence consisted of three parts: 
1. Instruction – A brief directive indicating the task and specifying which keys to track for value updates. 
2. Update stream – The input consists of a sequence of key–value pairs, where a fixed set of keys each receive an equal number of updates. The updates for different keys are randomly interleaved and are organized such that the same key does not appear in two consecutive key-value pairs. This setup mimics concurrent updates in real-world data logs, without grouping updates for the same key contiguously. 
3. Query – A prompt instructing the model to retrieve the final value associated with each tracked key.

The retrieval objective was to return the most recent value associated with each specified key. For each key with update count X, the preceding (X–1) key–value pairs served as irrelevant, interfering updates—sharing the same key but differing in value. This design allowed us to isolate the effect of interference. 

Because the retrieval target is always the value from the last occurrence of each key, search difficulty is inherently low: the model simply needs to locate the most recent update for each key. While the random interleaving of updates does not guarantee that the last occurrence of each key is near the end of the sequence, the retrieval target’s relative position is always clearly defined as the most recent appearance. As a result, the search space is small and well-defined. The main challenge, therefore, is not finding the target, but correctly identifying it in the presence of multiple earlier, competing updates for the same key. This mirrors realistic data environments where many variables are updated concurrently, and interference—rather than search—becomes the limiting factor.

In this particular experiment, we used 46 unique keys, each receiving multiple value updates throughout the sequence. For each key, the last value it receives is the retrieval target, while all earlier key–value pairs for that key—totaling 46 × (update count – 1) interfering distractors—serve to induce interference. (Figure\ref{fig:input_updates}) provides an example input and its corresponding output for three keys undergoing multiple updates. We measure accuracy by counting the number of correctly retrieved final values across all keys.

\noindent\textbf{Data and Performance Evaluation} \\
To maintain comparability with human performance, we constructed a word dictionary with up to 46 categories, each comprising 400 words. The token lengths of words within each category were selected to fall within a similar range. Keys were drawn from these category names, and values were randomly selected from the corresponding categories in the dictionary. This dictionary design aligns with cognitive psychology proactive interference tests related to human working memory. Words were randomly selected from the dictionary in each test run to eliminate the potential effects of specific semantic combinations. Confidence intervals (CI95) were computed using bootstrap methods after multiple test repetitions.

This synthetic key–value retrieval task is closely related to “Lost-in-the-Middle” \citep{liu-liang2024LostMiddleHow}, which examines how the position of the retrieval target within the context affects accuracy. In contrast, our approach offers finer experimental control over interference: by always probing the most recently updated value for each key, we hold the target’s relative position constant. In later experiments, we also fix the total input length, allowing us to systematically isolate and measure the effects of interference in the retrieval task.

\noindent\textbf{Models} \\
We evaluated a broad spectrum of state-of-the-art open-source and proprietary LLMs, ranging from 0.6B (Qwen3-0.6B) to 637B parameters (Deepseek-V3), and including major proprietary models such as GPT, Claude, Gemini, and Grok. Our benchmark covers both dense and Mixture-of-Experts (MoE) architectures, spanning diverse training data volumes and hardware resources.

\subsection{Results and Discussion}

\begin{figure*}[ht]
  \centering
  \includegraphics[width=0.80\linewidth]{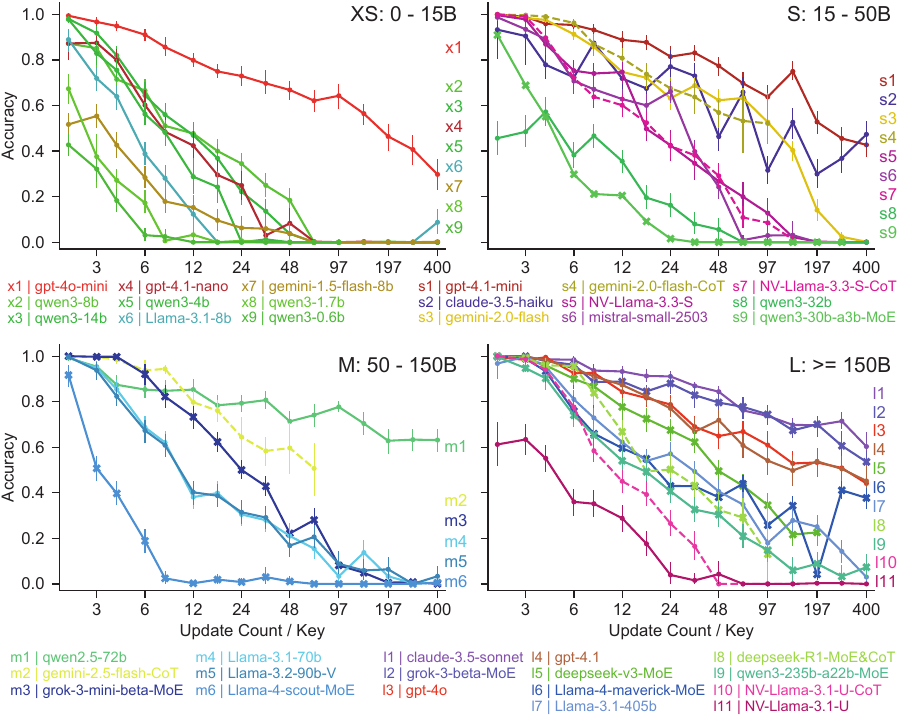}
\caption{Model retrieval accuracy declined approximately log-linearly as the number of updates per key increased. For visualization, models were grouped by estimated parameter size into four tiers—XS, S, M, and L—shown from top to bottom. Larger models (L group) tended to degrade more slowly, while smaller models (XS group) declined the fastest. The x-axis is log-scaled, covering update counts from 3 to 400 (log-spaced). The number of keys updated was fixed at 46. Most models exhibited a quasi-log-linear drop, reaching near-zero performance midway or projected to do so at higher update counts. Models were color-coded by developer, with similar hues representing models from the same organization. Each panel (e.g., for "M" models) assigns sequential labels (m1, m2, m3, ...) to models, following the IES ranking shown in Figure~\ref{fig:ies_rank}. If applicable, model names were suffixed with “MoE” or “CoT”. On the plots, MoE models are indicated with “X” markers, and CoT variants with dashed lines. These color, style, and naming conventions are used consistently across figures. Error bars indicate bootstrapped 95\% confidence intervals.}
  \label{fig:acc_updates}
\end{figure*}

In this first experiment, we investigated how different levels of interference affected the model’s ability to retrieve information—specifically, the most recent value associated with each key. To manipulate interference strength, we fixed the number of unique keys at 46 and varied the update count—the number of key–value pairs presented per key—from 3 to 400. A higher update count corresponded to a greater interference load for each queried key. By requiring the model to retrieve only the most recently presented value of each key, we kept search difficulty low and isolated the effect of interference. This design allowed us to systematically assess how increasing the amount of interfering information impacts retrieval accuracy.

\textbf{Interference information significantly impairs the ability of LLMs to effectively utilize context information.} Across models of varying parameter sizes, we observe a robust log-linear decline in retrieval accuracy as additional interfering key–value pairs are inserted before the target value for each key (Figure~\ref{fig:acc_updates}). This log-linear trend reflects rapid initial accuracy loss, with subsequent interference causing smaller additional declines. Notably, the log-linear effect persists across models of different developmental stages and model sizes; larger models exhibit a more gradual decline than smaller ones. For example, both M and L models begin with perfect accuracy, but at 400 updates per key, the L models maintain relatively high accuracy, whereas the M group’s accuracy drops sharply to 0\%---with only one exception—indicating a near---complete inability to retrieve the most recent value at the end of the context window.

\textbf{Robustness to Prompt Variations}

Although absolute retrieval accuracy can shift with changes in prompt wording \citep{he-hasan2024DoesPromptFormatting}, our study emphasizes the relative trend of performance decline rather than raw accuracy scores. This approach effectively neutralizes variability arising from individual prompt formulations.

To further confirm the robustness of the observed proactive interference (PI) effect, we tested additional prompt templates explicitly designed to verify task comprehension. Specifically, we introduced meta-relevant prompts that first ask the LLM to articulate the "task mission"—for example, explicitly prompting the model to "describe the goal of this task" before retrieval. This step ensured the models fully understood the retrieval objective of identifying "the last value" (see Figure~\ref{fig:acc_forget_reduced}, ‘Relevance meta-prompt’). Across these prompt variations, the qualitative trend of performance decline—specifically, the log-linear decay in accuracy—remained consistently robust.

\begin{figure}[h]
  \centering
  \includegraphics[width=1\linewidth]{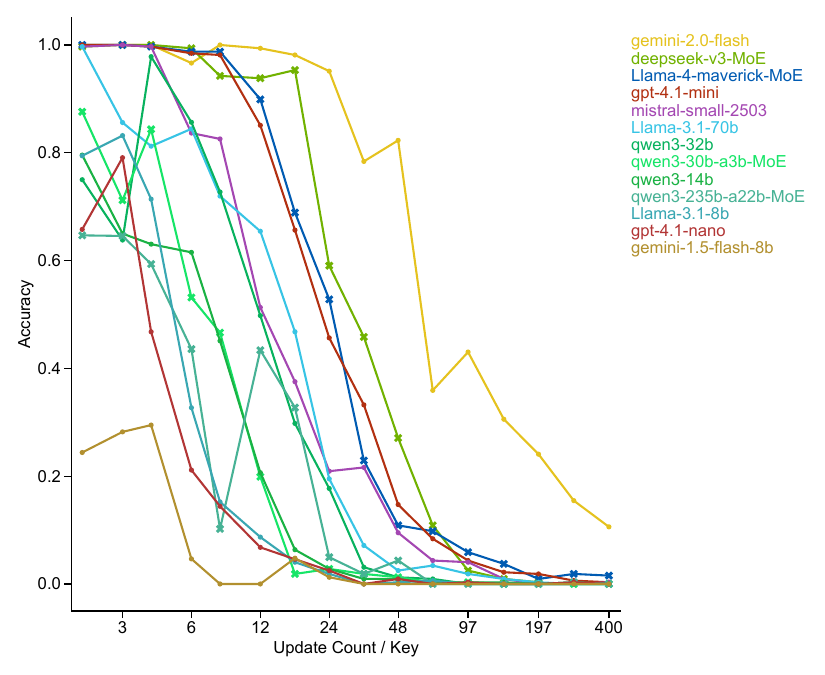}
  \caption{
  \textbf{Step-like failure pattern in sequential key–value update tests.}
  Retrieval accuracy remains near-perfect as interfering information is added in strictly sequential order, until a model-specific threshold is reached—after which performance drops abruptly to near-zero. Within the same model family, larger models exhibit a higher threshold (better capacity). Despite quantitative differences, all models show the same two-plateau, step-function pattern, reflecting a hard capacity limit. This stands in contrast to the gradual log-linear decay observed under random update order (see Fig.~\ref{fig:acc_demo}). (x-axis: number of interfering items, log-scaled; asterisk: MoE models)
  }
  \label{fig:acc_seq_demo}
\end{figure}

Additionally, we rearranged the input organization to test PI under both randomly shuffled and strictly sequential update sequences (i.e., sequential key–value updates without randomization; see Figure~\ref{fig:acc_seq_demo}). Notably, in sequential mode, retrieval accuracy remains stable until reaching a model-specific interference threshold, after which performance sharply and consistently drops to near-zero—a two-plateau, step-like pattern contrasting with the gradual log-linear decay observed in random mode. The consistent interference-induced decline across diverse models and input structures further underscores the robustness and generalizability of the observed PI phenomenon.

\subsubsection{\textbf{Incorrect extractions are primarily attributed to proactive interference}}

Given the consistent decline in LLM extraction accuracy when interference information is introduced, we investigated the underlying causes of these errors. Our analysis of the input sequences that appeared in the LLM's responses reveals that \textbf{errors are predominantly influenced by information encountered before the final, correct update} to a given key. This phenomenon is analogous to proactive interference (PI) in cognitive science, where previously learned information hinders the retrieval of more recent information.

We observe a three-stage progression in error distribution patterns as interference increases:

\textbf{Stage 1 – Low Interference, Tightly Focused Errors}:
When interference is low, retrieval accuracy is high and the model’s error distribution is sharply peaked around the correct value. Errors, when they occur, are not random but show a consistent pattern: they tend to be earlier key–value pairs for the same key, typically located in positions (bins) immediately preceding the final, correct value.This indicates that the model’s confusion is narrowly constrained and spatially localized.

\textbf{Stage 2 – Moderate Interference, Dispersed Errors:}:
As interference increases, retrieval accuracy drops, and the output distribution spreads. Retrieval errors now stem from much earlier updates—far upstream from the target value rather than adjacent positions (bins). 
with a small but growing fraction now involve values never presented at all (“hallucinations”). This increasing dispersion marks rising proactive interference and a decline in retrieval fidelity.

\textbf{Stage 3 – High Interference, Hallucinatory Responses}  
At high levels of interference, retrieval accuracy collapses and the model’s output distribution undergoes a qualitative shift. The model increasingly returns values that never appeared in the prompt—so-called hallucinations. At the same time, a substantial portion of errors remains anchored to the earliest bins, reflecting a persistent primacy bias toward the first few updates for each key, even as retrieval fidelity breaks down. This change in retrieval behavior resembles a phase transition: once the model’s anti-interference capacity is exhausted, it no longer retrieves plausible candidates, consistent with limited-resource theories of working memory failure.

\begin{figure}[h]
  \centering
  \includegraphics[width=1\linewidth]{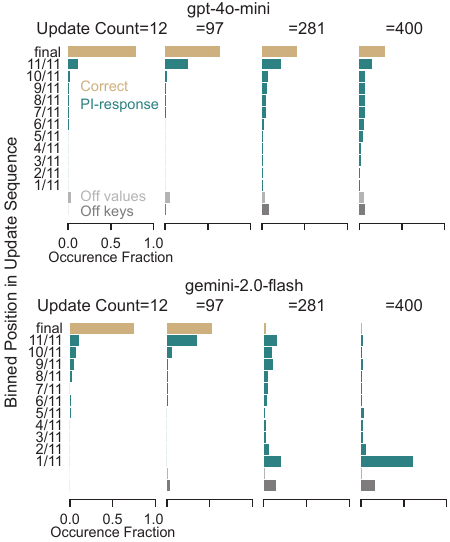}
  \caption{Distribution of model responses across update positions, showing increasing signs of PI as update count increases (left to right). The y-axis lists 11 equal-width bins (Bin 1–Bin 11, green) covering the entire update sequence. The earthy yellow bar indicates the single final update—the correct retrieval target. Light gray bars (“off values”) denote cases where the model returns a value not present in the update history (i.e., hallucinations). Dark gray bars (“off keys”) indicate failures to return any value for the queried key. As update count increases, errors shift from clustering near the final update to earlier bins, with rising rates of off-values and off-keys. For response distributions from additional models, see Figure~\ref{fig:resp_updates} in the Appendix.}
  \label{fig:resp_updates_reduced}
\end{figure}

Figure~\ref{fig:resp_updates_reduced} illustrates this progression: as the update count increases (moving left to right in the panels), the model’s incorrect responses shift from the most recent value to much earlier, outdated values, and eventually to off-target ‘hallucinated’ values.

This distribution change aligns with our "limited resource" observation, suggesting a phase transition in LLM behavior once their anti-interference resources are depleted. In human studies, PI resilience is highly correlated with working memory capacity. Our results suggest that an LLM's anti-interference capability could serve as a metric for its working memory capacity—not just its ability to store information, but to actively maneuver and manage it.

To strengthen the generalizability of our findings, we conducted additional experiments on a broader set of models, with consistent results shown in Supplementary Figure~\ref{fig:resp_updates}

\subsubsection{\textbf{Size Over Input Context Length}}
Statistical tests confirm that anti-interference performance correlates with model size and is weakly correlated with the context window.

To quantify each model’s robustness to interference, we introduce the Interference Endurance Score (IES). The IES is defined as the area under the curve (AUC) of retrieval accuracy, calculated across log-scaled update counts. We measure how well a model maintains accurate retrieval as interference increases, with a higher IES indicating greater resistance to interference. For comparability and statistical reliability, we compute the IES using the accuracy-versus-update-count function (see Figure 5), which is available for all models tested.  

To determine whether model performance is driven more by parameter size or by context length, we conducted a regression analysis of the Interference Endurance Score (IES) against both variables. We grouped models into four parameter size classes—XS, S, M, and L—as defined in Figure~\ref{fig:acc_updates}. Reasoning models were excluded because their more extensive inference processes caused latency to exceed 200 seconds per task, preventing most tests from completing. To minimize noise from closed-source models with uncertain parameter counts, we focused on these defined size classes and restricted our analysis to open models.

\begin{figure}[h]
  \centering
  \includegraphics[width=0.9\linewidth]{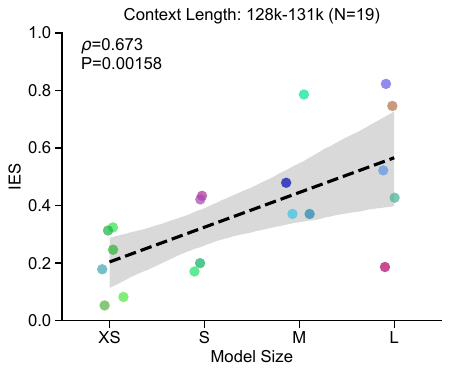}
  \caption{Interference Endurance Score (IES, from Figure~\ref{fig:ies_family}) shows a strong correlation with model size class (XS, S, M, L; as defined in Figure~\ref{fig:acc_updates}). Each dot represents a model, color-coded as in Figure~\ref{fig:ies_family}. A linear regression line is included for visualization, with shaded regions indicating 95\% confidence intervals. The analysis is restricted to models with similar context lengths (128k–131k tokens, covering about two-thirds of tested non-CoT models. R-squared value is derived from Spearman correlation.}
  \label{fig:regression}
\end{figure}

The regression shows that parameter size class is a significant predictor of IES (t = 3.03, p = 0.005, N = 30), while context length has no significant effect (t = –0.144, p = 0.886). The combined model explains 26.1\% of the variance in IES (R² = 0.261). To further clarify the role of model size, we performed a separate analysis restricted to models with similar context lengths (128k–131k tokens), which encompasses two-thirds of the non-CoT models. Within this range, the Spearman correlation between parameter size and IES remains strong and significant ($\rho^2 = 0.673$, p = 0.0016; see Figure~\ref{fig:regression}).
.

Our analysis shows that

\textbf{Model size—not context window length—is the primary factor that underlies robustness to interference.}

\textbf{MoE architectures underperform dense models with comparable total parameters} (we conjecture that this is because the number of activated parameters in an MoE model is much smaller than its nominal total).

\begin{figure}[ht]
  \centering
  \includegraphics[width=0.9\linewidth]{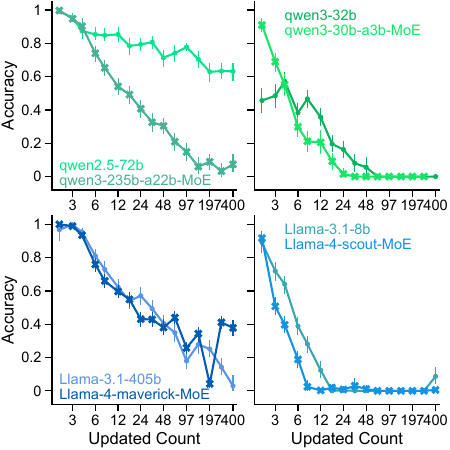} 
  \caption{Comparison of retrieval accuracy between Mixture-of-Experts (MoE) and dense models.
Each curve shows retrieval accuracy versus update count for a single model. MoE models are denoted by "X" markers and labeled with “MoE” in the legend. Across update counts, MoE architectures consistently match or underperform dense models with similar total parameter counts, and in many cases perform comparably to much smaller dense models.
(MoE models shown: Llama-4-maverick-MoE (400B), Llama-4-scout-MoE (109B), Qwen3-30B-A3B-MoE (30B).)}
  \label{fig:acc_moe}
\end{figure}

In cognitive science, performance under proactive interference is a classic probe of working-memory capacity: individuals with greater ability to maintain and manipulate information show greater resistance to interference. Our findings reveal a striking parallel in large language models (LLMs). Across all tested LLMs, we observe a consistent, characteristic decline in retrieval accuracy as interference increases. Moreover, larger models demonstrate greater resistance to interference—a pattern reminiscent of individual differences in human working memory.

This universal decline, present even in state-of-the-art models spanning a wide range of scales, training data, and computational resources, suggests that limited resistance to interference is an inherent property of transformer-based architectures, rather than a byproduct of specific model size or dataset.

Importantly, our metric captures more than just the context window length or the sheer amount of information a model can store. It quantifies each model’s effective ability to manage and control information in the presence of substantial distractors—tracking, updating, and selectively retrieving relevant data amid interference. Thus, anti-interference performance reflects not only storage capacity, but also the executive control processes that underlie working memory in humans. This framework enables us to operationalize and compare the working-memory-like functions of LLMs and human cognition on a principled, quantitative basis.

\section{Interference Is Independent of Input Length}

Retrieval accuracy in language models declines log-linearly as the update count per key increases, suggesting a limited working-memory-like capacity. However, in the previous experiment, input length was not controlled; thus, the observed decline might simply reflect increasing context length rather than genuine interference. To directly test the role of interference, we designed two additional manipulations.
\begin{enumerate}
    \item \textbf{Number of Updated Keys}: Increasing the number of distinct keys that are updated within the context, while holding the update count per key constant.
    \item \textbf{Partial Query at Fixed Input Length}: Fixing both the total number of keys and the update count per key (thus keeping the input length constant), but varying the number of keys queried—asking the language model to track and retrieve only a subset of the keys presented.
\end{enumerate}

By manipulating interference both with and without changes in input length, we can dissociate the effects of interference from those of context length; observing similar declines in retrieval accuracy across both manipulations would provide strong evidence that interference, rather than context length alone, constrains model performance.

\subsection{Experiment Setup}

\subsubsection{\textbf{Experiment A: Varying the Number of Updated Keys}}

In this experiment, we fixed the update count for each key (either 125 or 350 update count per key), and systematically increased interference by varying the number of distinct keys presented in the sequence-the Updated Keys (U, from 1 to 46). This contrasts with our earlier experiment, which held the number of keys constant while varying the update count per key. 

For each input sequence, there are U relevant key–value pairs, with the retrieval target being the last value for each key. Depending on the update count, this results in U × (125 – 1) or U × (350 – 1) irrelevant, interfering key–value pairs per input sequence. Retrieval accuracy was measured as a function of U-Updated Keys .

\subsubsection{\textbf{Experiment B: Fixed Length Version}}

To further isolate the effect of interference, we designed a complementary experiment in which the total input length was held constant. In this condition, both the update count per key and the number of updated keys (U) were fixed, so each input sequence contained the same number of key–value pairs. However, we varied the number of keys the model was instructed to track and retrieve at the end—these are the Tracked Keys (T), chosen from among the U updated keys.

Specifically, the update count for each key was fixed (either 125 or 350 updates per key), and the number of updated keys U was also fixed at 46. We then systematically varied the number of tracked keys (T, from 1 to 46), i.e., the subset of keys for which the model was asked to report the final value. Retrieval accuracy was measured as a function of T, the number of tracked keys.

\begin{figure}[h]
  \centering
  \includegraphics[width=1\linewidth]{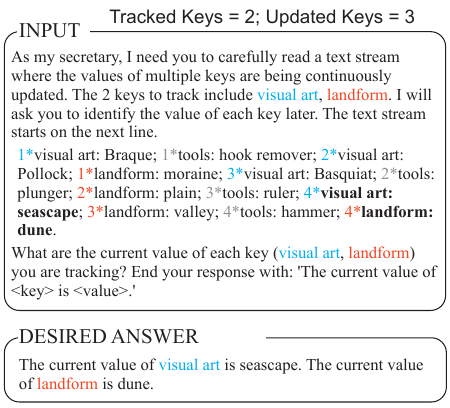}
  \caption{Input example illustrating how the model is prompted to track and return values for a subset of updated keys, as specified by the parameters tracked keys and updated keys. In this minimal example, the tracked keys include "visual art" (blue) and "landform" (orange); the "tools" key (prefixed with a gray index like “1*”) appears in the update stream but is not referenced in the initial instruction or final query. Ideally, the model should return only the most recent values for the tracked keys. This setup enables testing whether model performance depends primarily on task-relevant information, rather than irrelevant updates or input length. Bold text highlights the target key-value pairs the model is expected to retrieve.}
  \label{fig:input_keys}
\end{figure}

Figure~\ref{fig:input_keys} provides an example input: among 3 distinct keys updated in the sequence, only 2 are tracked (queried) at the end.

\begin{figure*}[h]
  \centering
  \includegraphics[width=0.8\linewidth]{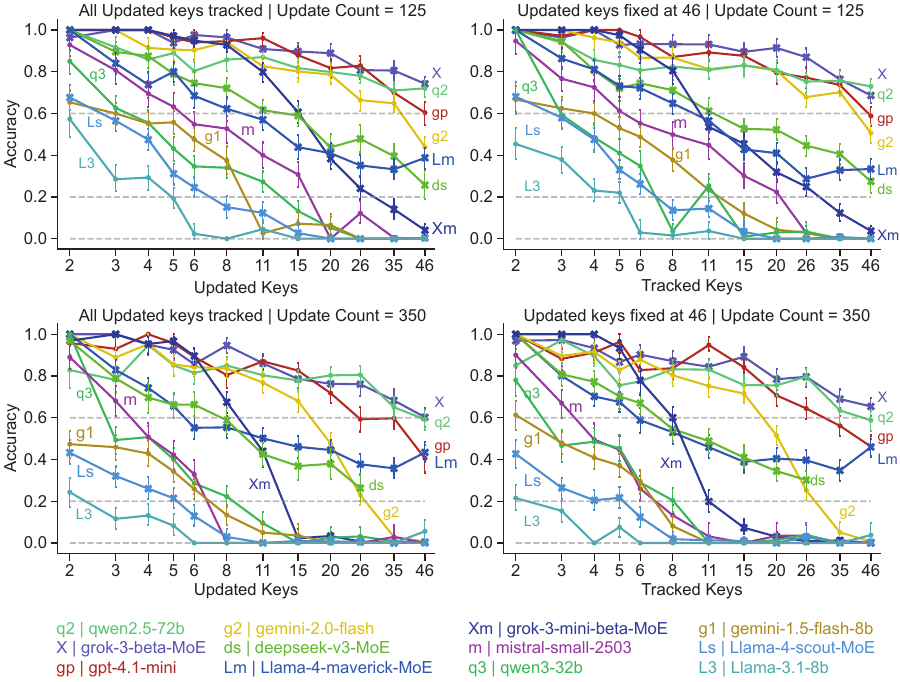}
  \caption{Varying the number of updated keys (left panels) versus the number of tracked keys (right panels, with updated keys fixed at maximum) yields only minor differences in retrieval accuracy. In all conditions, accuracy declines approximately log-linearly with the number of keys. Each key is updated a fixed number of times—125 in the upper panels and 350 in the lower panels. Some models exhibit a two-phase decline; for example, grok-3-mini-beta maintains high performance early on, followed by a sharp drop after a turning point. deepseek-v3 does not complete the full range in the lower panel due to context length limitations. MoE models are indicated by "X" markers. Error bars represent 95\% confidence intervals computed via bootstrapping. Model acronyms are used to label the corresponding curves.}
  \label{fig:acc_keys_reduced}
\end{figure*}

\subsection{Result Figures for Both Experiments}

\subsubsection{Experiment A result}

Increasing interference by raising the number of updated keys consistently produced a log-linear decline in retrieval performance across all tested model sizes (Left Panel of Figure~\ref{fig:acc_keys_reduced} ). Notably, even though each key received a fixed number of updates—ensuring a constant interference load per key—requiring the model to retrieve the final values for a greater number of keys more rapidly exhausted its anti-interference resources, leading to a substantial reduction in accuracy.

Specifically, in the Left Panel of Figure~\ref{fig:acc_keys_reduced}, the x-axis represents the total number of Updated Keys, and models are instructed to track all Updated Keys. Each key's update count is fixed at two values: 125 (upper panel) and 350 (lower panel). The overall trend in log-scale is a linear decline in accuracy, independent of the number of updates per key. 

\subsubsection{Experiment B fix length result}

Retrieval performance exhibits a consistent log-linear decline across all tested models (Right Panel of Figure~\ref{fig:acc_keys_reduced} ). The x-axis represents the total number of Tracked Keys.  Notably, larger models show shallower declines than smaller ones, reflecting greater resistance to interference. Under fixed input length, increasing the number of simultaneously tracked keys leads to lower accuracy, in line with this log-linear pattern. For instance, llama4-maverick achieves nearly 100\% accuracy when tracking just two keys, but this drops below 5\% when tracking 46 keys, consistently following the same downward trajectory. These results indicate that, under fixed-length conditions, tracked keys compete for a limited pool of anti-interference resources, which are rapidly depleted as their number grows. Practically, this suggests that reducing the number of concurrently tracked keys can substantially improve retrieval accuracy.

\subsubsection{Combined Observations (Experiments A and B)}
We observe that both Experiments A and B exhibit nearly identical log-linear declines in retrieval accuracy as the number of tracked keys increases (see Figure~\ref{fig:acc_keys_reduced}, left and right panels). Notably, this occurs even though Experiment B keeps input length fixed while Experiment A allows it to grow. This similarity indicates that the observed performance drop cannot be attributed solely to longer input sequences; rather, it is driven by increased interference from tracking more keys.

Furthermore, models that excel in the variable input length setting (Experiment A) also perform well in the fixed-length setting (Experiment B), underscoring the robustness of this pattern across experimental setups.

\textbf{The universal log-linear decline observed, even under fixed input length, suggests that anti-interference capacity operates as a distinct resource, separate from the total context window length.} In other words, regardless of how much context the model can technically process, its ability to manage interference is independently limited. This distinction highlights that interference resistance is a fundamental capability of LLMs—determined not by context window size, but by deeper architectural or computational constraints within the model.

\subsubsection{Discussion/Implications}
These findings have important implications for both model evaluation and practical deployment. They suggest that simply increasing the context window or scaling up input length does not directly translate into better interference management. Instead, targeted advances in anti-interference mechanisms or executive control within model architectures may be needed to substantially improve retrieval accuracy when handling many competing, similar items. This perspective reframes interference resistance as a critical axis of model capability, worthy of focused research and explicit benchmarking alongside traditional context-length and parameter-count metrics.

\section{Retrieval Capacity Is Limited by a Single Interference Bottleneck Across Dimensions}

If an LLM’s anti-interference capacity is truly analogous to human working memory, then manipulations that increase working memory demands in humans should produce comparable effects in LLMs. One such manipulation is the classic word-length effect: in human memory research, increasing the length of words to be remembered impairs performance, as longer items consume more working memory resources  \citep{baddeley1975word}. This phenomenon provides an additional axis along which working memory capacity can be taxed.

\begin{figure}[h]
  \centering
  \includegraphics[width=0.95\linewidth]{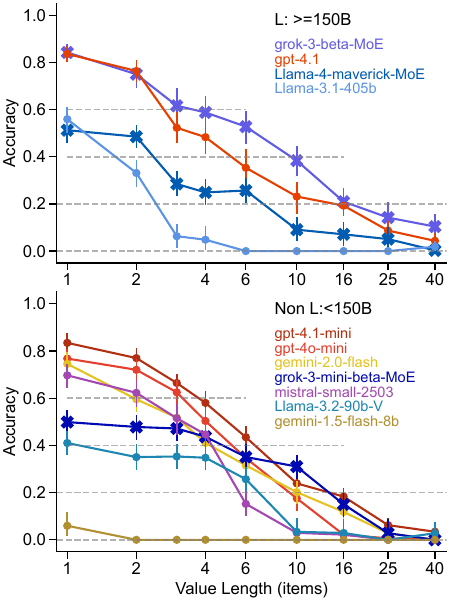}
  \caption{Retrieval accuracy as a function of value length, showing a roughly log-linear decline toward near-zero performance. For clarity, models are grouped by parameter size: large models (L; $\geq$150B parameters) are shown in the upper panel, and smaller models ($<$ 150B) in the lower panel. The update count is fixed at 20. Some models exhibit an initial plateau phase, with stable accuracy for short value lengths (ranging from 1 to 4). At the largest value length tested, accuracy drops to near zero for most models, with the exception of Grok-3-beta, which retains a performance of approximately 0.1. MoE models are indicated with “X” markers. Error bars represent bootstrapped 95\% confidence intervals.}
  \label{fig:acc_value_length}
\end{figure}

To probe whether LLMs exhibit a similar sensitivity, we systematically varied the length of words within key–value pairs by concatenating multiple words into each value. This allowed us to directly test whether increasing the information load per item would similarly degrade retrieval performance in LLMs.

In this experiment, we held constant the three previously identified sources of interference: the number of updates per key, the number of updated keys, and the number of keys to track. To manipulate interference strength in line with the classic word-length effect observed in human working memory, we systematically increased the length of the updated value strings. Specifically, we concatenated multiple dictionary words end-to-end (e.g., \textbf{AppleOrangeBanana}), thereby \textbf{increasing both the word length and the token count—the fundamental unit of LLM processing.} This manipulation closely mirrors the increased cognitive load humans experience when encoding longer words in memory tasks. Figure~\ref{fig:input_value_length} in the supplement provides an example input.

\subsection{Results and Interpretation}
\textbf{LLMs exhibit a universal, approximately log-linear decline in retrieval accuracy as the length of each value increases.} The slope of this decline is markedly steeper than for the other three interference manipulations: increasing value length from one to ten words drives accuracy below 40\% for every model tested, and extending it to forty words reduces accuracy to under 5\%. Notably, this sharp drop occurs even when the number of keys, updates, and tracked keys is held constant, highlighting the unique impact of item length.

This result demonstrates that increasing the amount of information stored in each retrieved value—by concatenating more words—adds a distinct, independent dimension of interference, taxing the system’s capacity beyond what can be explained by the number of tracked keys or updates alone. The effect of value length thus exposes another axis along which the model’s anti-interference resource can be depleted.

This outcome closely parallels human memory performance, where recalling longer or more complex words substantially lowers accuracy—a classic word-length effect. Taken together with prior results, these findings reinforce our explanatory framework: all forms of interference—whether from more keys, more updates, or longer values—tap into a single, \textbf{unified anti-interference resource in the model, analogous to a working-memory buffer.} As the informational load per item grows, this capacity is consumed more rapidly, leading to steeper performance degradation. This unified capacity constraint, shared across all tested dimensions, underscores a structural limitation in current LLM architectures that mirrors properties of human working memory.

\section{Mitigating Interference: Empirical Insights from LLM–Human Comparison}

\begin{figure*}[ht]
  \centering
  \includegraphics[width=0.85\linewidth]{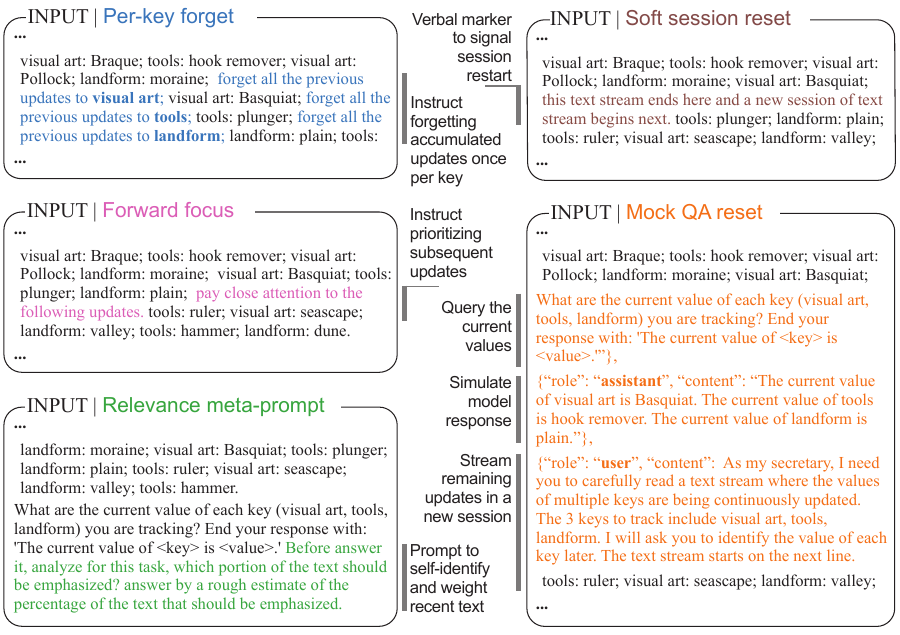} 
  \caption{Example input illustrating intervention strategies designed to mitigate proactive interference. Each strategy inserts explicit cues into the update stream, typically near the end (e.g., at the 120th-last update or one-third before the final update). The five strategies are: \textbf{Per-key forget} (green): An instruction to disregard previous updates for a specific key before a new one (e.g., "forget all the previous updates to visual art"). \textbf{Forward focus} (magenta): An instruction to prioritize information that follows (e.g., "pay close attention to the following updates"). \textbf{Relevance meta-prompt} (green): A prompt for the model to self-assess and estimate the proportion of text to prioritize before answering. \textbf{Soft session reset} (brown): A verbal cue marking the start of a new input segment (e.g., "this text stream ends here and a new session of text stream begins next"). \textbf{Mock QA reset} (orange): A simulated dialogue turn including an initial update segment, a query, a mock assistant response, and remaining updates in a new user turn. Inserted instructional cues are shown in colored text; role labels in the Mock QA reset are in bold.}
  \label{fig:input_forget}
\end{figure*}

Our previous experiments demonstrated that LLMs possess limited anti-interference capacity, with retrieval accuracy declining log-linearly as interference increases. To better understand and potentially mitigate this limitation, we compared LLM performance to humans on the same key-value retrieval task, drawing on strategies from human cognitive experiments to design corresponding interventions for LLMs.

In contrast to LLMs, humans exhibit a plateau in recall accuracy for the most recent key-value pair, even as the number of prior updates accumulates. Classic working memory studies attribute this resilience to executive control mechanisms. Two particularly relevant mechanisms are gating, which automatically suppresses or discards outdated information as new items are encoded \citep{oberauer-vockenberg2009UpdatingWorkingMemory}, and directed forgetting, where individuals intentionally discard certain information when explicitly instructed to do so \citep{festini-reuter-lorenz2014CognitiveControlFamiliarity}.

LLM retrieval accuracy declines continuously with increasing interference, suggesting the absence of automatic gating mechanisms. Moreover, humans can engage in explicit, strategic forgetting. To test whether LLMs might benefit from such strategic forgetting, we simulated this human capability by providing LLMs explicit prompts instructing them to forget previous key-value pairs. If successful, such an intervention would demonstrate that LLMs can emulate human-like release from interference through external cues, potentially alleviating the limitations of their anti-interference capacity.

\subsection{Simulating Human Directed Forgetting with Natural Language Prompt}
To simulate the human strategy of explicit directed forgetting, we inserted a targeted prompt into the input sequence that directly instructs the LLM to disregard all prior updates for a specific key. This prompt is placed at a fixed point in the update stream—immediately before the chunk containing the majority of the target answer, after most interfering updates have been presented. The directive reads: “Forget all the previous updates to {key},” with {key} dynamically replaced by the relevant key for the current task.

The purpose of this intervention is to actively suppress the influence of outdated or distracting information from earlier in the sequence, thereby reducing proactive interference and guiding the model to prioritize only the most recent updates for retrieval. This approach tests whether an explicit natural language cue can shift the model’s focus in a way that mimics human executive control over memory. See Figure~\ref{fig:input_forget} for examples of the "Per-Key Forget" prompt.

For comparison, we also tested a "Forward Focus" prompt, which instructs the LLM to concentrate on the more recent, relevant portion of the input. This allows us to evaluate whether explicit natural language instructions—whether aimed at forgetting or focusing—can meaningfully affect model retrieval performance.

\begin{figure}[h]
  \centering
  \includegraphics[width=0.95\linewidth]{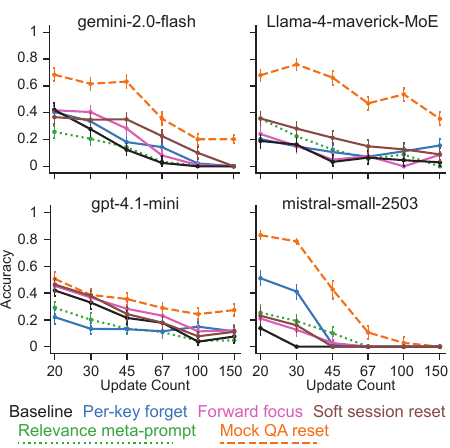}
  \caption{Explicit forgetting and focusing prompts inserted during the update stream (as shown in Figure~\ref{fig:input_forget} yielded only marginal improvements in retrieval accuracy. \textbf{The black line indicates the baseline} condition with no intervention prompt. \textbf{Solid lines} represent several simple \textbf{natural language prompts} designed to instruct the model to forget previous updates, focus on upcoming ones, or reset context. For most models, these \textbf{interventions had limited effect}, especially at higher update counts, where the baseline performance is low. \textbf{The per-key forget(blue line)} even had a negative effect on gpt-4.1-mini. The \textbf{relevance meta-prompt (green dotted)}, which asked the model to self-assess what to focus on, was ineffective for all models and even harmful for gpt-4.1-mini. \textbf{Only the mock QA reset intervention (orange dashed line)}, which simulates a user-model interaction, led to a \textbf{substantial improvement} in retrieval accuracy. However, this \textbf{strategy was not immune to the overall trend: accuracy continues to decline with increasing update count (log-spaced).}Experiments used 46 unique keys and a key-value pair length of 6.(Additional results on other models, confirming this trend, are provided in Supplementary Figure~\ref{fig:acc_forget}.)}
  \label{fig:acc_forget_reduced}
\end{figure}

\subsection{Natural Language Prompt Fails}

The per-key forget prompt—designed to mimic human explicit forgetting—did not significantly improve retrieval accuracy (blue line in Figure~\ref{fig:acc_forget_reduced}; $\Delta < 10$ percentage points compared to the baseline at 100 updates, black line). Similarly, alternative natural language instructions intended to focus the model on the target retrieval section were also ineffective. Overall, \textbf{natural language instructions—whether to forget or focus—do not effectively reduce interference in LLM retrieval tasks.}

\begin{figure}[h]
  \centering
  \includegraphics[width=1\linewidth]{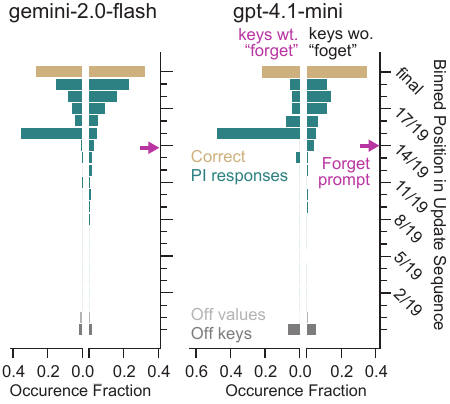}
  \caption{A selective per-key forgetting prompt induces a distinct pattern of Proactive Interference (PI): instead of enabling successful forgetting, the prompt causes retrieval errors to cluster around the position in the update sequence where the instruction was injected. The figure compares keys that received a forgetting instruction with a control group; only the instructed keys show this pronounced error concentration, indicating that the prompt anchored the model's retrieval errors to that part of the sequence rather than erasing the information. Earthy yellow bars indicate the correct value—the final update. Green bars represent earlier (interfering) values, grouped by their relative position in the update sequence. Light gray bars show “off values” not present in the update history, and dark gray bars denote “off keys,” where the model failed to return any value. The results shown are from an experiment with 20 updates per key and 46 unique keys. For a comprehensive analysis across various model architectures and a wider range of parameter settings, see Figure~\ref{fig:resp_half_forget} in the Appendix.}
  \label{fig:resp_half_forget_reduced}
\end{figure}

An analysis of the error distribution reveals a critical failure mode: rather than improving retrieval accuracy, the per-key forgetting prompt consistently caused errors to cluster around the position in the sequence where the instruction was injected. As shown in Figure~\ref{fig:resp_half_forget_reduced}, models displayed a pronounced tendency to select earlier values immediately preceding the forget instruction, rather than the correct, final update. This error pattern indicates that the prompt did not enable the model to effectively disregard prior information. Instead, it induced a concentration of retrieval errors near the instructed forget position, reshaping interference rather than mitigating it. In summary, \textbf{rather than mitigating interference, these prompts cause errors to cluster around the location of the prompt, indicating that the model’s anti-interference limitation cannot be overcome by simple natural language cues.}

\subsection{Hack method: Mock-QA-reset succeeds}

Inspired by LLM “hacking” studies showing that models can be coaxed to bypass earlier instructions \citep{kuo-chen2025HCoTHijackingChainofThought}, we devised a non-natural-language mock QA reset prompt (see Figure~\ref{fig:input_forget}) that mimics human directed-forgetting. Inserted 120 updates before the final query, this reset cue leads the model to treat preceding input as belonging to an already processed prior task, thereby partially mitigating interference and improving retrieval accuracy. While this ad hoc prompt intervention partially reduces interference, it highlights the need for more systematic methods to address interference in LLMs' retrieval task.

The prompt has three parts:
\begin{itemize}
  \item \textbf{Simulated user query} asking for the current value of all tracked keys  
        (e.g., ``User: What is the current value of key1, key2, …, key45?''), which frames prior updates as a closed batch.
  \item \textbf{Simulated assistant reply} giving fabricated answers  
        (e.g., ``Assistant: The current value of key1 is …, key2 is …''), providing explicit closure.
  \item \textbf{New user prompt} signalling a fresh tracking task, followed by the remaining updates  
        (e.g., ``User: I will now provide 45 updated key–value pairs. Tell me the most current value. The pairs begin:''), which marks a clear task boundary and encourages the model to ignore earlier content.
\end{itemize}
This artificial task boundary partially mitigates interference by prompting the model to deprioritize earlier input and focus on newly updated information.

\textbf{The hack prompt substantially improved retrieval accuracy (As shown in Figure~\ref{fig:acc_forget_reduced}, orange line), reducing the effects of interference across all tested LLMs.} This hack-based prompt consistently outperformed natural language instructions designed to induce forgetting or refocusing. For example, with the hack-reset, Gemini Flash 2.0’s retrieval accuracy at 150 update count—under high interference—matched its baseline performance at only 30–45 update count, demonstrating a substantial reduction in interference effects. 

The success of our hacking-based reset method demonstrates that implementing a gating mechanism can effectively reduce interference in LLMs, closely mirroring the executive gating functions of human working memory. This result suggests that implementing gating mechanisms in LLMs could be an effective strategy for reducing interference, mirroring the executive functions of human working memory.

However, while our reset strategy shows that LLMs benefit from artificially imposed context boundaries, this approach remains fundamentally limited. Specifically, our intervention mitigates interference by diminishing the influence of all prior information—effectively discarding or bypassing past associations. Although this provides short-term relief, it is not a viable solution for real-world tasks, which frequently require selective, context-dependent access to historical data beyond just the most recent update.

This limitation is further highlighted by additional experiments with natural-language prompts designed to instruct LLMs to ignore or forget prior information—such as the 'soft session reset' shown in Figure~\ref{fig:input_forget}—which were largely ineffective (see performance in Figure ~\ref{fig:acc_forget_reduced}). These findings indicate that current LLMs cannot be reliably controlled through explicit natural-language user instructions alone; precise, natural-language-based adjustments of memory and attention remain an open challenge.

\section{Top down vs Bottom up: Why CoT does not improve information retrieval under interference}

\begin{figure}[h]
  \centering
  \includegraphics[width=1\linewidth]{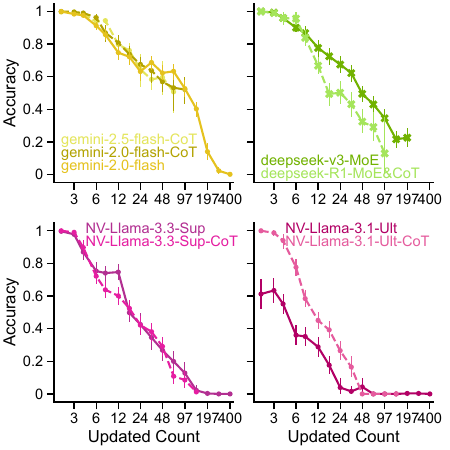} 
  \caption{Chain-of-Thought (CoT) does not improve retrieval performance. Accuracy as a function of update count is shown for four pairs of CoT-enabled models and their corresponding base (non-CoT) versions. In three of the four comparisons, the CoT variant performs worse than or equal to its base model. Upper left panel: gemini-2.0-flash-thinking matches its base model (within its tested range). The newer gemini-2.5-flash (CoT) performs identically to gemini-2.0-flash-thinking. Upper-right panel: deepseek-R1 (CoT) is outperformed by the base model–deepseek-V3 beyond update count 12. lower-left panel: nvidia-llama-3.3-nemotron-super and its CoT variant perform equally well. Lower-right panel: nvidia-llama-3.1-nemotron-ultra is the weakest base model, and the only case where CoT improves performance—though still below nvidia-llama-3.3-nemotron-super. CoT variants are only tested up to about 100 updates or less due to output overflow ("thinking" exhaustion). Solid lines denote base models; dashed lines denote CoT variants. The x-axis is log-scaled.}
  \label{fig:acc_cot}
\end{figure}

Additionally, when we compared various prompt styles, we found that LLMs were susceptible to interference across all tested prompts. To eliminate potential ambiguities in task instruction, we designed an "activate-locate" prompt (full prompt in Relevance meta-prompt in Figure~\ref{fig:input_forget}), which first required the LLM to analyze and state the location of the target key-value pair within the input. While models could correctly identify that the answer was at the very end, this knowledge did not translate into improved retrieval performance; accuracy remained comparable to baseline conditions and still exhibited a consistent decline. (Relevance meta-prompt in Figure~\ref{fig:acc_forget_reduced})  

This indicates a discrepancy between an LLM's top-down analytical capabilities and its bottom-up information processing and retrieval behavior: knowing \emph{where} the answer is does not necessarily improve its ability to retrieve it under interference. This result provides a potential explanation for why CoT models do not improve information retrieval under interference.

To further substantiate this finding, we conducted additional tests across a range of models and retrieval settings. We evaluated various models alongside their CoT-enabled counterparts, including open-source models (Deepseek V3, Deepseek R1), proprietary models (Gemini Flash 2.0, Gemini Flash 2.5 with CoT), and state-of-the-art models offering both CoT and non-CoT modes (such as Nvidia-Llama). Consistently, we observed that the latency and cost of CoT models are significantly higher than those of non-CoT models on this retrieval task, yet CoT does not improve performance.

\section{Conclusion}

Our systematic investigation of proactive interference (PI) in Large Language Models (LLMs) across various scales from 0.6B to over 600B, reveals a pervasive susceptibility to interference effects during retrieval tasks. Critically, LLMs demonstrate a continuous log-linear decline in retrieval accuracy as interference increases, showing no evidence of a plateau. Moreover, the continuous decline—characterized by a similar log-linear pattern—emerges independently along multiple dimensions of interference load: the number of sequential updates to a key, the number of keys tracked concurrently, and the token length of each updated value. The convergence of these qualitatively similar decline patterns, across orthogonal axes of load, implies that LLMs possess a finite, resource-like representational capacity that can be incrementally taxed by different, yet functionally interchangeable, forms of cognitive load, independent of total input size or the model’s maximum context length. Collectively, these findings indicate that the anti-interference capacity observed in LLMs closely parallels the properties of human working memory.

We also identify a critical dissociation between the analytical and execution capabilities of LLMs: even models capable of explicitly articulating effective retrieval strategies fail to translate this analytical understanding into improved retrieval performance, underscoring a lack of top-down executive control over retrieval tasks.

Our findings establish proactive interference as a pervasive failure mode in contemporary LLMs and introduce a novel interpretation: a model’s resistance to proactive interference directly reflects its underlying working-memory capacity. Unlike traditional metrics that emphasize total input length, our approach reveals interference resilience as a distinct, cognitively-grounded dimension of model capability. Since interference is inherent to tasks ranging from summarizing repeatedly updated information to conducting complex, long-horizon reasoning, enhancing LLMs’ working-memory robustness becomes critical for practical performance. By providing a structured synthetic evaluation framework explicitly designed to measure susceptibility to interference, this study offers both a diagnostic tool and a theoretical advance toward understanding and improving LLM cognition. Our code and datasets are publicly released to foster further investigation into the memory mechanisms of large language models.

\section{Notes}
All experiments were concluded by May 5th, 2025. For a detailed list of model versions, please refer to Appendix~\ref{sec:model-versions}.

\section{Discussion}

\subsection{Methodological Contribution}

\subsubsection{From Cognitive Paradigms to LLM Diagnostic Tools}
Our current work aligns squarely with this research thrust. We do not merely suggest that Proactive Interference (PI) is a problem for LLMs by analogy to humans; we adapt the specific experimental logic of the A-B, A-C, A-D paired-associate learning paradigm—a workhorse of human PI research—to create a novel, synthetic diagnostic tool for LLMs. This allows for controlled experimentation and the systematic manipulation of variables, moving beyond correlational observations from general benchmarks towards a more causal understanding of LLM failure modes. 

Crucially, the A-B, A-C (and, by extension, A-D, A-E…)  schema captures a vast class of real-world problems: streaming sensor readouts, mutable legal ledgers, and long reasoning chains in which the same variable is updated and queried repeatedly. By embedding this ubiquitous “value-overwriting” structure into our testbed, we ensure that the experiment speaks to both practical performance gaps and deeper theoretical questions about how LLMs process interfering information.

By manipulating specific variables known to affect a cognitive process (like PI) and observing the direct impact on LLM performance in a controlled setting, we can begin to build and test mechanistic hypotheses about LLM cognition and its points of failure. For example, if increasing the number of A-B presentations before A-C leads to stronger PI in LLMs (as it typically does in humans), this provides evidence that the LLM's representation of "learning strength" or "associative trace" behaves analogously to human working memory in the context of interference. This "cognitive probe" approach offers greater diagnostic precision for specific failure modes than is achievable with broad-stroke performance benchmarks.

\subsection{Theoretical explanations and implications}
Our results suggest that current LLMs possess only an implicit, resource-bounded form of memory selectivity.  Self-attention weights provide a quasi-executive filter that suffices for low–interference conditions, but unlike human prefrontal gating, it cannot be strengthened or re-allocated on demand. When the interference budget is exceeded, the model’s retrieval accuracy degrades monotonically to near-zero performance, revealing the absence of a true top-down control system. Because adaptive executive control over memory is widely held to be a core component of goal-directed intelligence, these findings point to a critical gap between contemporary LLMs and human cognition: transformers can store vast contexts, but they cannot decide how to use—or forget—them. 

\subsubsection{Connecting Behavioral Evidence with Mechanistic Interpretability}
Our research also complements work on the mechanistic interpretability of LLMs, such as the study of induction heads \citep{anthropic_incontext_2022}. While induction heads offer a plausible mechanism for how in-context learning and subsequent interference might occur (e.g., an induction circuit strongly encoding A-B might resist an A-C update), our paper provides the experimental paradigm to test the behavioral consequences of such mechanisms when they are confronted with conflicting associative information. Our synthetic setup is designed precisely to probe the conditions under which these induction-like mechanisms are robust versus when they are susceptible to PI.

Recent studies have applied human working-memory (WM) tests, such as the N-back paradigm, to assess the possible WM capacity of LLMs \citep{gong-wang2024WorkingMemoryCapacity}. While prior work has primarily focused on transplanting classic cognitive tests from human studies to LLMs, our approach integrates cognitive science methodologies with tasks modeled on realistic LLM applications. This design enables more ecologically valid assessments—reflecting typical model usage—and allows for direct comparison of LLM and human retrieval abilities on matched tasks. Utilizing proactive interference as a framework, we identify specific behavioral differences and practical limitations that standard benchmarks often fail to reveal. These results underscore the importance of integrating cognitive and applied perspectives to advance research on LLM capabilities.

\section*{Code Availability}
The code for all experiments is publicly available at: \url{https://github.com/zhuangziGiantfish/Unable-to-Forget}

\section*{Author Contributions}
*Equal contribution. Listing order is random.
Chupei Wang (CW) and Jiaqiu Vince Sun (JVS) co-initiated the development of cognitive-inspired working memory stress tests to reveal this limitation. CW proposed framing LLM retrieval errors as an interference problem and led the iterative trial-and-error testing. JVS implemented and automated the main experiment pipelines, and managed data collection and analysis.

CW designed Experiment 1 revealing how increasing interference degrades LLM retrieval accuracy. Both authors collaborated to engineer Experiment 2, carefully controlling input size to isolate the effects of interference. JVS expanded the framework with Experiment 3, showing that increasing value length intensifies interference. This indicates LLMs’ interference resistance as a unified capacity, exhaustible across different dimensions.

To address the deficits exposed, JVS developed a natural-language intervention and thoroughly analyzed prompt effects and error distributions, while CW pioneered a prompt-hacking strategy, drawing on detailed empirical observations of LLM behavior to mitigate interference.

\section*{Acknowledgments}
Special thanks to Sam Zheyang Zheng, Simons Foundation’s Flatiron Institute CCN researcher and NYU Ph.D. candidate, whose clear, contemplative discussions with Chupei(CW)—linking human and LLM focus to dynamical systems—encouraged the initiation of this investigation.

We also thank Yikun Kuang (Ph.D. student, Center for Data Science–CILVR Lab, NYU), whose consultations—from early brainstorming to conference advice—enriched our work by offering a machine learning perspective that complemented our cognitive approaches.

\section*{Impact Statement}
This paper presents work whose goal is to advance the field of Machine Learning. There are many potential societal consequences of our work, none of which we feel must be specifically highlighted here.


\bibliography{common/refs.bib}
\bibliographystyle{icml2024}

\newpage
\appendix
\section{Appendix: Detailed Graph}

\begin{figure}[h]
  \centering
  \includegraphics[width=1\linewidth]{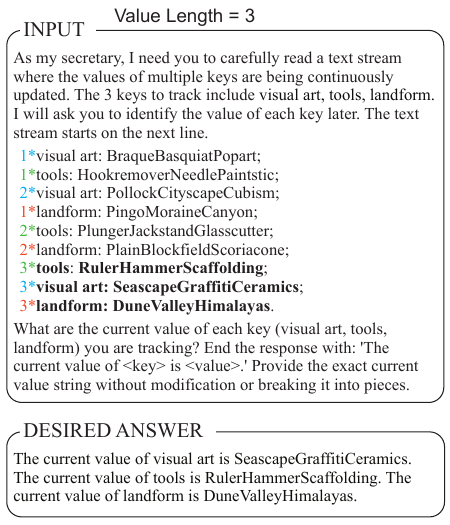}
  \caption{Input example with manipulation of the updated values's length. In this example, three items from the same category are space-removed, capitalized at the first letter, and concatenated into a single updated value. Bold text indicates the target key-value pairs the model is expected to retrieve.}
  \label{fig:input_value_length}
\end{figure}

\begin{figure}[h]
  \centering
  \includegraphics[width=1\linewidth]{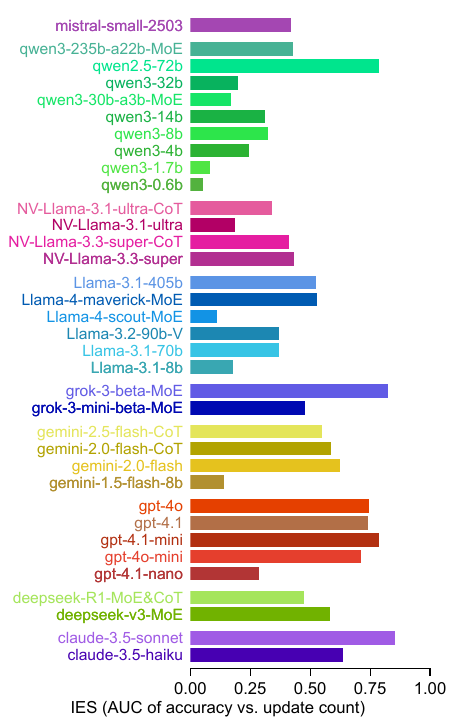}
  \caption{Interference Endurance Score (IES) for all models shown in Figure 3, computed as the area under the curve (AUC) of their accuracy–update-count functions in Figure~\ref{fig:acc_updates}. Higher IES indicates greater robustness to interference across increasing update counts. Models are grouped by family using the same color scheme as in Figure 3, and within each family, sorted by parameter size from large (top) to small (bottom). For a ranking of IES values by magnitude, see Figure~\ref{fig:ies_rank} in the Appendix}
  \label{fig:ies_family}
\end{figure}

\begin{figure}[h]
  \centering
  \includegraphics[width=\linewidth]{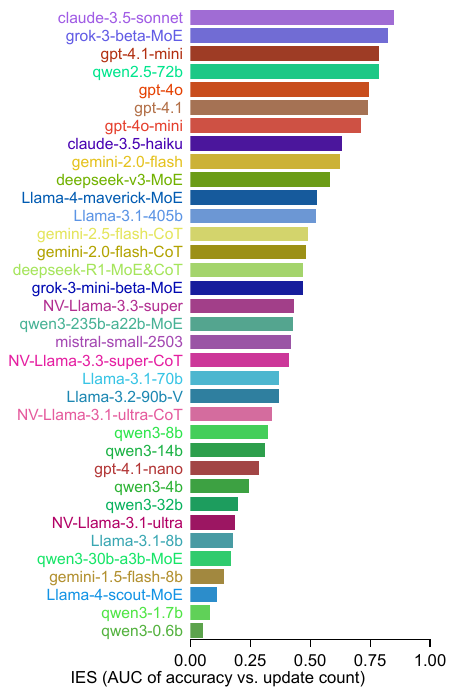} 
  \caption{Interference Endurance Scores (IES) from Figure~\ref{fig:ies_family}, re-ordered by IES value in descending order.}
  \label{fig:ies_rank}
\end{figure}

\begin{figure*}[h]
  \centering
  \includegraphics[width=0.95\linewidth]{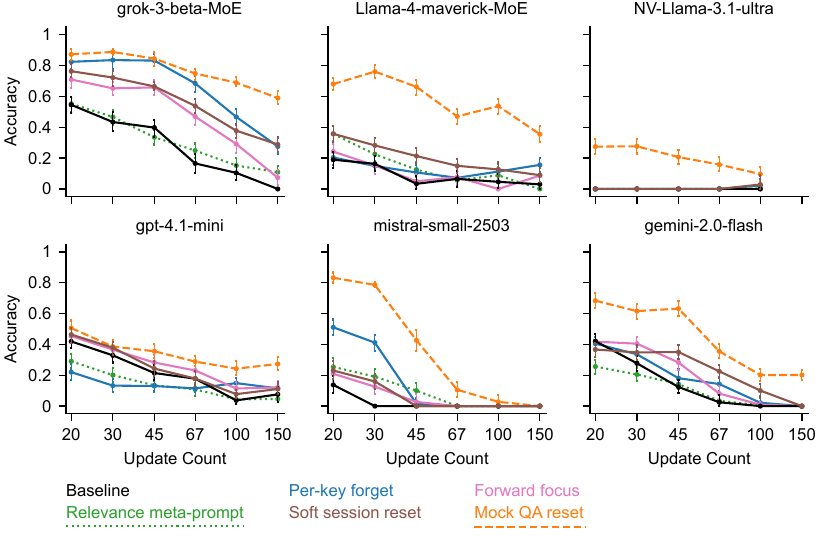}
  \caption{Explicit forgetting and focusing prompts inserted during the update stream (as shown in Figure~\ref{fig:input_forget} yielded only marginal improvements in retrieval accuracy. \textbf{The black line indicates the baseline} condition with no intervention prompt. \textbf{Solid lines} represent several simple \textbf{natural language prompts} designed to instruct the model to forget previous updates, focus on upcoming ones, or reset context. For most models, these \textbf{interventions had limited effect}, especially at higher update counts, where the baseline performance is low. \textbf{The per-key forget(blue line)} even had a negative effect on gpt-4.1-mini. The \textbf{relevance meta-prompt (green dotted)}, which asked the model to self-assess what to focus on, was ineffective for all models and even harmful for gpt-4.1-mini. \textbf{Only the mock QA reset intervention (orange dashed line)}, which simulates a user-model interaction, led to a \textbf{substantial improvement} in retrieval accuracy. However, this \textbf{strategy was not immune to the overall trend: accuracy continues to decline with increasing update count (log-spaced).}Experiments used 46 unique keys and a key-value pair length of 6.}
  \label{fig:acc_forget}
\end{figure*}

\begin{figure*}[h]
  \centering
  \includegraphics[width=0.9\linewidth]{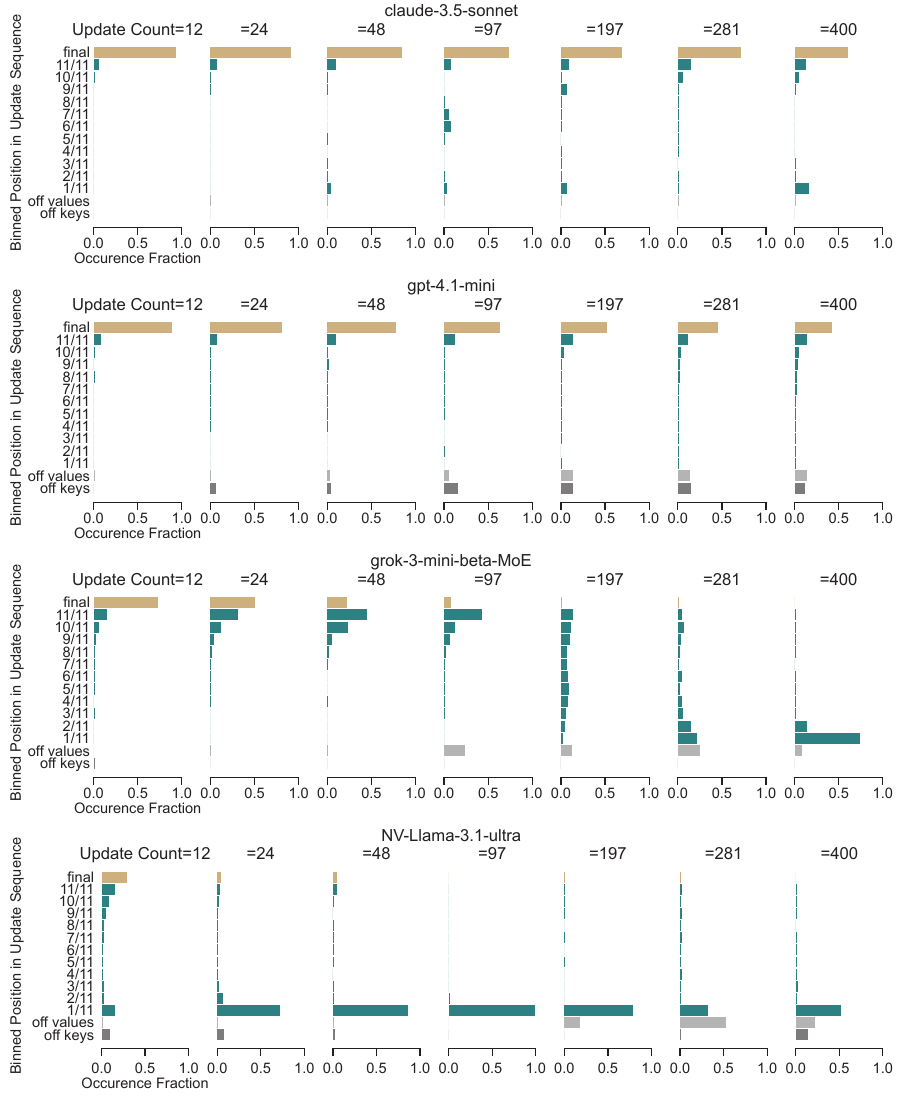} 
  \caption{Distribution of model responses across update positions, showing increasing signs of PI as update count increases (left to right). The y-axis lists 11 equal-width bins (Bin 1–Bin 11, green) covering the entire update sequence. The earthy yellow bar indicates the single final update—the correct retrieval target. Light gray bars (“off values”) denote cases where the model returns a value not present in the update history (i.e., hallucinations). Dark gray bars (“off keys”) indicate failures to return any value for the queried key. As update count increases, errors shift from clustering near the final update to earlier bins, with rising rates of off-values and off-keys. See Figure~\ref{fig:resp_updates} for additional models}
  \label{fig:resp_updates}
\end{figure*}

\begin{figure*}[h]
  \centering
  \includegraphics[width=0.77\linewidth]{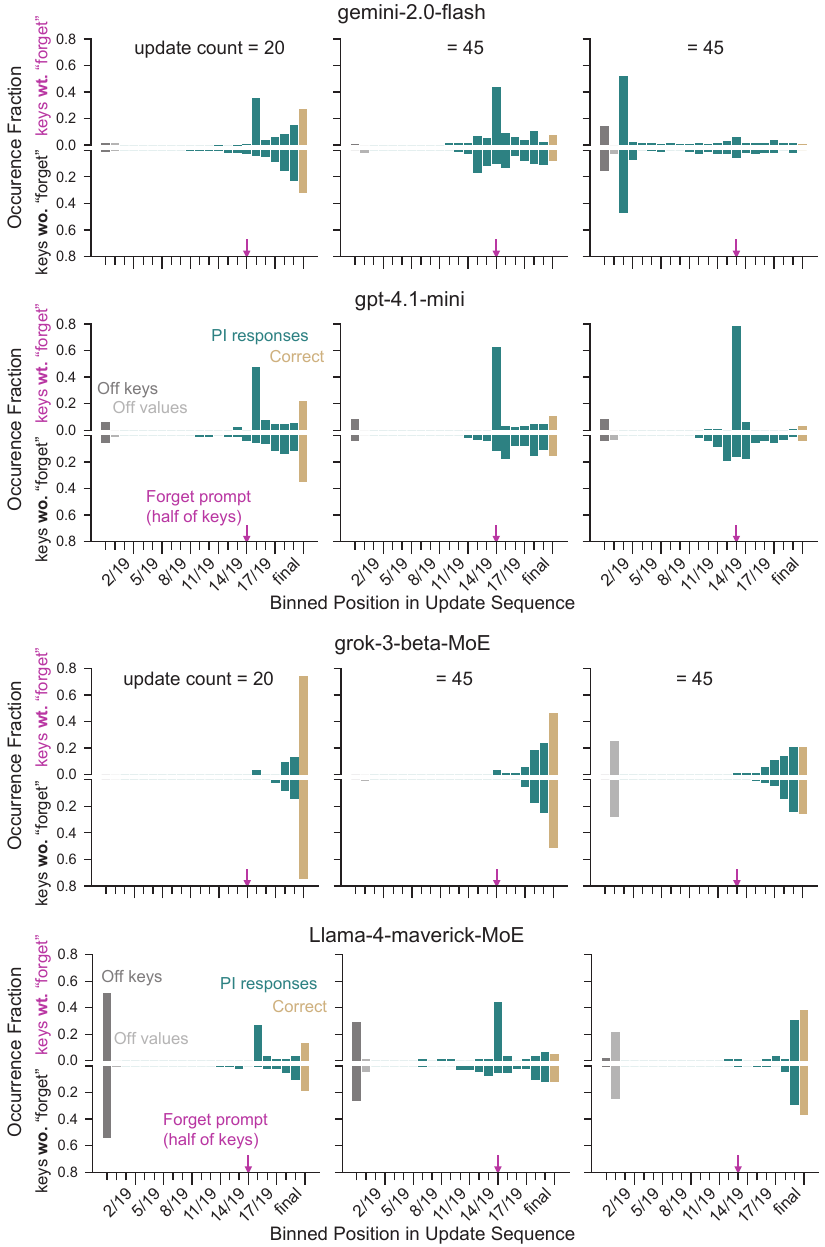}
  \caption{The selective per-key forgetting prompt amplifies proactive interference. Keys that received a forgetting instruction prior to the final third of their updates exhibited concentrated errors around the forgetting point, compared to keys without such a prompt. The x-axis indicates the position of the selected value within the update sequence, categorized for each key. Earthy yellow bars indicate the correct value—the final update. Green bars represent earlier (interfering) values, grouped into 19 bins based on their relative position in the update sequence. Light gray bars indicate “off values” not present in the update history. Dark gray bars denote “off keys,” where the model failed to return any value.}
  \label{fig:resp_half_forget}
\end{figure*}

\onecolumn 

\section{Model Versions}
\label{sec:model-versions}

This section provides a comprehensive list of all language models used in our evaluation. All experiments were conducted up to May 5th, 2025. Models with explicit date stamps in their identifiers (e.g., \texttt{gpt-4o-2024-11-20}) represent fixed snapshots. For other models, we used the versions detailed below.

\begin{longtable}{ll}
\toprule
\textbf{Model Name} & \textbf{Version/Snapshot} \\
\midrule
\endfirsthead

\multicolumn{2}{c}%
{{\bfseries \tablename\ \thetable{} -- continued from previous page}} \\
\toprule
\textbf{Model Name} & \textbf{Version/Snapshot} \\
\midrule
\endhead

\midrule
\multicolumn{2}{r}{{Continued on next page}} \\
\endfoot

\bottomrule
\endlastfoot

\multicolumn{2}{l}{\textit{Google Gemini Models}} \\
\midrule
gemini-2.5-flash-preview & gemini-2.5-flash-preview-04-17 \\
gemini-2.0-flash & gemini-2.0-flash \\
gemini-1.5-flash-8b & gemini-1.5-flash-8b \\
gemini-2.0-flash-thinking-exp\footnotemark[2] & gemini-2.0-flash-thinking-exp-01-21 \\
\midrule

\multicolumn{2}{l}{\textit{OpenAI Models}} \\
\midrule
gpt-4.1 & gpt-4.1-2025-04-14 \\
gpt-4.1-mini & gpt-4.1-mini-2025-04-14 \\
gpt-4.1-nano & gpt-4.1-nano-2025-04-14 \\
gpt-4o & gpt-4o-2024-11-20 \\
gpt-4o-mini & gpt-4o-mini-2024-07-18 \\
\midrule

\multicolumn{2}{l}{\textit{DeepSeek Models}} \\
\midrule
deepseek-chat & deepseek-chat\footnotemark[1] \\
deepseek-reasoner & deepseek-reasoner\footnotemark[1] \\
\midrule

\multicolumn{2}{l}{\textit{Alibaba Qwen Models}} \\
\midrule
qwen2.5-72b-instruct & qwen2.5-72b-instruct\footnotemark[1] \\
qwen3-0.6b & qwen3-0.6b\footnotemark[1] \\
qwen3-1.7b & qwen3-1.7b\footnotemark[1] \\
qwen3-4b & qwen3-4b\footnotemark[1] \\
qwen3-8b & qwen3-8b\footnotemark[1] \\
qwen3-14b & qwen3-14b\footnotemark[1] \\
qwen3-32b & qwen3-32b\footnotemark[1] \\
qwen3-30b-a3b & qwen3-30b-a3b\footnotemark[1] \\
qwen3-235b-a22b & qwen3-235b-a22b\footnotemark[1] \\
qwen3-0.6b-thinking\footnotemark[2] & qwen3-0.6b-thinking\footnotemark[1] \\
qwen3-1.7b-thinking\footnotemark[2] & qwen3-1.7b-thinking\footnotemark[1] \\
qwen3-4b-thinking\footnotemark[2] & qwen3-4b-thinking\footnotemark[1] \\
qwen3-8b-thinking\footnotemark[2] & qwen3-8b-thinking\footnotemark[1] \\
qwen3-14b-thinking\footnotemark[2] & qwen3-14b-thinking\footnotemark[1] \\
qwen3-32b-thinking\footnotemark[2] & qwen3-32b-thinking\footnotemark[1] \\
qwen3-30b-a3b-thinking\footnotemark[2] & qwen3-30b-a3b-thinking\footnotemark[1] \\
qwen3-235b-a22b-thinking\footnotemark[2] & qwen3-235b-a22b-thinking\footnotemark[1] \\
\midrule

\multicolumn{2}{l}{\textit{Meta LLaMA Models}} \\
\midrule
llama-4-maverick-17b-128e-instruct-maas & llama-4-maverick-17b-128e-instruct-maas\footnotemark[1] \\
llama-4-scout-17b-16e-instruct-maas & llama-4-scout-17b-16e-instruct-maas\footnotemark[1] \\
llama-3.1-405b-instruct-maas & llama-3.1-405b-instruct-maas\footnotemark[1] \\
llama-3.2-90b-vision-instruct-maas & llama-3.2-90b-vision-instruct-maas\footnotemark[1] \\
llama-3.1-70b-instruct-maas & llama-3.1-70b-instruct-maas\footnotemark[1] \\
llama-3.1-8b-instruct-maas & llama-3.1-8b-instruct-maas\footnotemark[1] \\
\midrule

\multicolumn{2}{l}{\textit{xAI Grok Models}} \\
\midrule
grok-3-beta & grok-3-beta\footnotemark[1] \\
grok-3-mini-beta & grok-3-mini-beta\footnotemark[1] \\
\midrule

\multicolumn{2}{l}{\textit{Anthropic Claude Models}} \\
\midrule
claude-3-5-sonnet & claude-3-5-sonnet-20241022 \\
claude-3-5-haiku & claude-3-5-haiku-20241022 \\
\midrule

\multicolumn{2}{l}{\textit{Mistral Models}} \\
\midrule
mistral-small-2503 & mistral-small-2503\footnotemark[1] \\
\midrule

\multicolumn{2}{l}{\textit{NVIDIA Models}} \\
\midrule
nvidia\_llama-3.1-nemotron-ultra-253b-v1 & nvidia\_llama-3.1-nemotron-ultra-253b-v1\footnotemark[1] \\
nvidia\_llama-3.3-nemotron-super-49b-v1 & nvidia\_llama-3.3-nemotron-super-49b-v1\footnotemark[1] \\
nvidia\_llama-3.1-nemotron-nano-8b-v1 & nvidia\_llama-3.1-nemotron-nano-8b-v1\footnotemark[1] \\

\end{longtable}

\footnotetext[1]{Model accessed via API. For models without explicit date suffixes, the identifiers listed represent the latest available API endpoints as of our testing cutoff date of May 5, 2025.}
\footnotetext[2]{The `-thinking` suffix indicates the model was evaluated with its native reasoning or Chain-of-Thought (CoT) mode enabled. Its corresponding counterpart without the suffix was evaluated with this mode disabled.}


\end{document}